\documentclass[10pt,twocolumn,letterpaper]{article}

\usepackage{iccv}
\usepackage{times}
\usepackage{epsfig}
\usepackage{graphicx}
\usepackage{amsmath}
\usepackage{amssymb}

\usepackage{caption}
\usepackage{subcaption}
\usepackage{booktabs}
\usepackage{balance}
\usepackage[dvipsnames,table]{xcolor}
\usepackage[percent]{overpic}
\usepackage{siunitx}
\usepackage{etoolbox}
\usepackage{pifont}

\usepackage[pagebackref=true,breaklinks=true,letterpaper=true,colorlinks,bookmarks=false]{hyperref}

\newcommand{\cmark}{\ding{51}}%
\newcommand{\xmark}{ }%

\iccvfinalcopy 

\def\iccvPaperID{6506} 

\ificcvfinal\fi

\begin{document}

\title{FACSIMILE:\\Fast and Accurate Scans From an Image in Less Than a Second}

\author{
    David Smith \qquad Matthew Loper \qquad Xiaochen Hu \qquad Paris Mavroidis \qquad Javier Romero \\
    Amazon Body Labs \\
    {\tt\small \{dlsmith,mloper,sonnyh,parism,javier\}@amazon.com}
}

\twocolumn[{%
\renewcommand\twocolumn[1][]{#1}%
\maketitle
\begin{center}
    \newcommand{\teaserwidth}{1.\textwidth}
    \vspace{-0.2in}
    \centerline{
    \begin{overpic}[trim={0cm .5cm 0cm 1.3cm},clip,height=5cm]{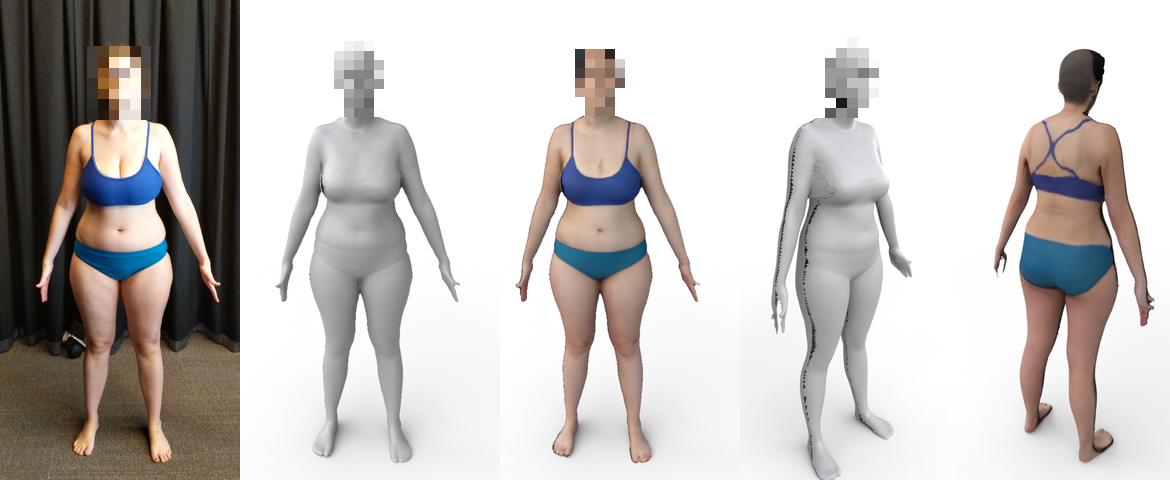}
      \put(1.0,.7){{\small \color{Melon} a}}
      \put(21.3,.7){{\small \color{gray} b}}
      \put(43.5,.7){{\small \color{Melon} c}}
      \put(64,.7){{\small \color{gray} d}}
      \put(82.5,.7){{\small \color{Melon} e}}
    \end{overpic}
    }
\vspace{-0.1in}
    \captionof{figure}{FAX converts a single RGB image (a) into a scan (b, d) with
    albedo texture (c, e)}
        \label{fig:teaser}
\end{center}%
}]

\maketitle
\ificcvfinal\thispagestyle{empty}\fi

\begin{abstract}
Current methods for body shape estimation either lack detail or require many images.
They are usually architecturally complex and computationally expensive.
We propose FACSIMILE (FAX), a method that estimates a detailed body from a single photo,
lowering the bar for creating virtual representations of humans.
Our approach is easy to implement and fast to execute, making it easily deployable.
FAX uses an image-translation network which recovers geometry at
the original resolution of the image. Counterintuitively, the main loss which drives
FAX is on per-pixel surface normals instead of per-pixel depth,
making it possible to estimate detailed body geometry without any depth supervision.
We evaluate our approach both qualitatively and quantitatively, and compare with a
state-of-the-art method.

\end{abstract}

\section{Introduction}

\begin{figure*}[htb]
    \centering
    \includegraphics[trim={0cm .5cm 1cm .4cm},clip,width=\linewidth]{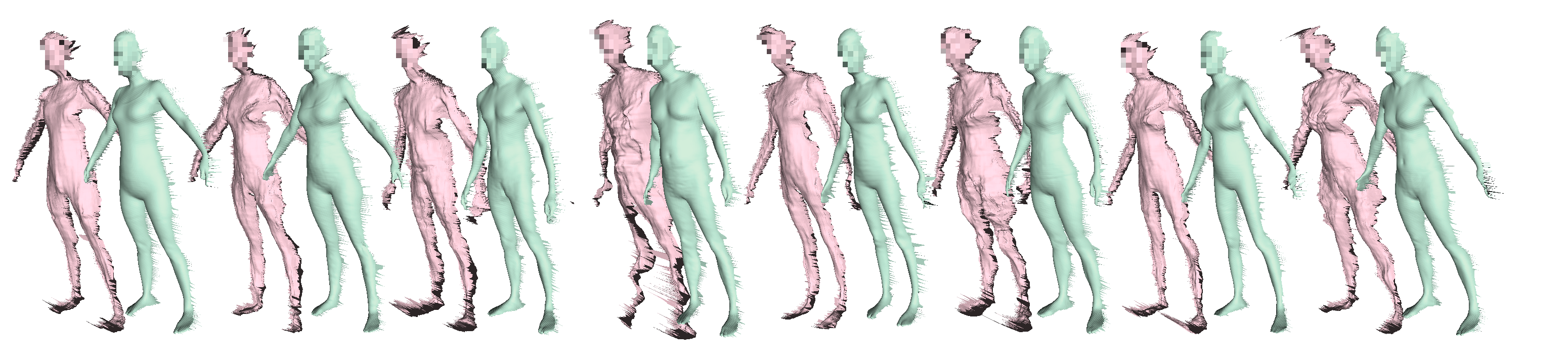}
    \caption{{
        Frontal meshes estimated using ({\bf pink}) an $L1$ loss on depth and ({\bf
        green}) an $L1$ loss on normals.
      }
    }
    \label{fig:d_l1_vs_n_l1}
\end{figure*}

High resolution body capture has not seen widespread adoption, despite a myriad of applications in medicine, gaming, and shopping. Traditional methods for high-quality body estimation require expensive capture systems which are difficult to deploy~\cite{Li13,bogo17dynamicfaust}.
More affordable RGB-D sensors like kinect have tried to overcome this problem~\cite{zhang14,Bogo2015detailed}, though those sensors are not as widespread as RGB cameras.
On the other hand, modern systems for single-photo body estimation lack detail~\cite{Dibra2017,siclope,alldieck2018detailed,kanazawa2018end,bogo2016keep,omran2018neural}. Our work is designed to help close the gap between an easily acquired image and a rich, detailed, reposeable avatar.

Systems targetted to recover shape from single images do a laudable job at recovering intermediate body representations. These include voxel-based reconstruction in~\cite{varol18bodynet}, the synthetic-view generation system in~\cite{siclope}, or the cross-modal neural nets in~\cite{Dibra2017}. But inevitably, the fidelity of their capture is limited by the granularity of their representation.

To address this lack of representational power, we apply modern image-to-image translation techniques~\cite{isola2017image,wang2017high} to geometry estimation. More concretely, we would like to estimate the depth corresponding to every foreground pixel in the image.
But this presents a new problem: the naive estimation of depth via an image translation network creates noisy, unusable surfaces (Figure~\ref{fig:d_l1_vs_n_l1}). This teaches us that when estimating depth with image-to-image translation, a direct loss on depth fails to give us a plausible surface.

The solution to this problem can be traced all the way back to Shape From Shading (SFS) literature by Horn~\cite{horn1986variational}, in which surface normals play a critical role in defining the
relationship between a surface and its appearance. Work focused in the reconstruction of the face region~\cite{richardson17learningcvpr} has shown that a loss on depth can benefit from an additional loss on
normals. We go beyond this insight showing that a loss just on normals can be \textit{sufficient} to reconstruct a high-quality depth map up to scale, and that this applies for an articulated, far from spherical object.

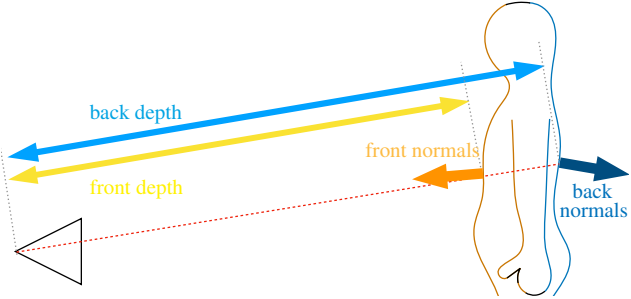
\begin{figure}[t!]
    \centering
    \begin{overpic}[trim={1.8cm 10.8cm 0.8cm 0.6cm},clip,width=\linewidth]{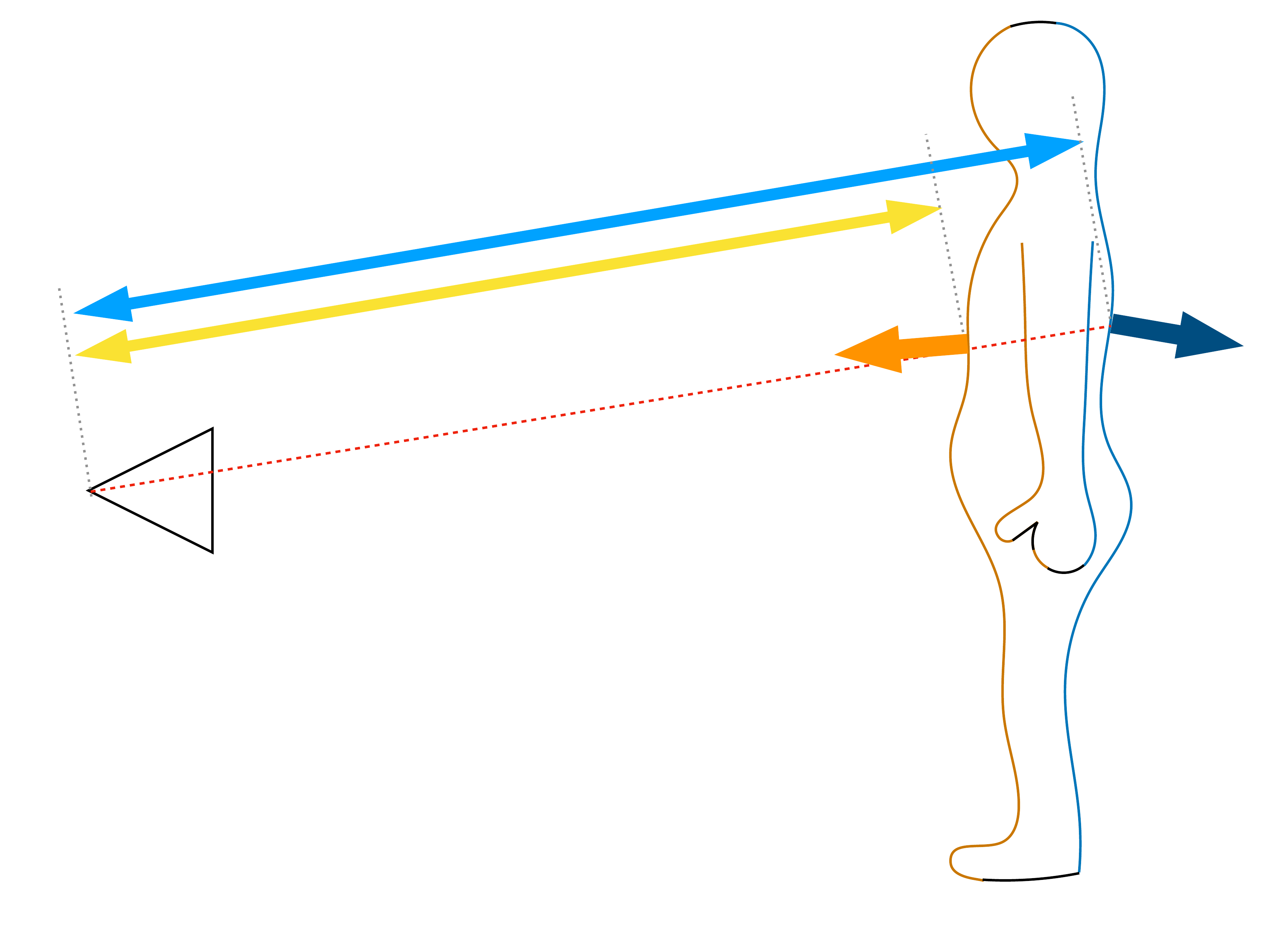}
      \put(14,16.1){{\footnotesize \color{Yellow} front depth}}
      \put(58,22.1){{\footnotesize \color{Dandelion} front normals}}
      \put(14,28.1){{\footnotesize \color{Cerulean} back depth}}
      \put(91,15.5){{\footnotesize \color{RoyalBlue} back }}
      \put(89,12.5){{\footnotesize \color{RoyalBlue} normals}}
    \end{overpic}
    \caption{
      Depth, surface normals
      and albedo are computed for the body points closest and furthest from the camera along
      the optic ray entailed by each pixel.
      Note the presence of pixels which remain unobserved and become holes in the
      inferred scan (black contours in the image).
    }
    \label{fig:silhouette}
    \vspace{-2mm}
\end{figure}

Because a single depthmap is still far from an entire avatar, we extended the system to estimate front and back-facing geometry and albedo. Similar to the concurrent work in~\cite{siclope}, we exploit the idea of obtaining two values per pixel by training the network to hypothesize the back side of the person (see Figure~\ref{fig:silhouette}). Unlike~\cite{siclope},
we do not restrict ourselves to texture and also estimate the back depth and normals. While current detailed methods like~\cite{siclope,alldieck2018detailed} typically take several minutes to run, we compute an almost complete scan containing geometry and texture in less than one second. In this publication we assume a \textit{cooperative subject} and focus on a specific type of image
that maximizes information capture (frontal arms-down pose, minimal clothing), although we believe the method could be applied to other cases and will continue investigating them in future work.

We demonstrate three contributions.
First, we compute full scans from a single image, orders of magnitude faster than current methods producing detailed scans.
Although other methods also reproduce garments, our method extracts significantly more detail. We encourage the reader to review the scans in figures~\ref{fig:teaser} and \ref{fig:seven_examples} and the supplementary material, paying special attention to subtle folds and compression artifacts in the chest, waist or hips, not present in any other methods.
Second, we show how these scans can be converted into detailed deformable avatars with little additional time (less than 10 seconds), which can be valuable for applications like gaming, measurements from an image, and virtual telepresence.
Finally, we illustrate the efficacy of our method
by comparing it quantitatively against the state-of-the-art multi-image method~\cite{alldieck2018video}
and performing a qualititative and quantitative ablation study.

\section{Related Work}

\textbf{Geometry estimation from a single-photo} has been a topic of research for at least 50 years.
Classic methods like \emph{shape from shading}~\cite{Horn1970}
take shading images and produce the underlying geometry.
Modern solutions to this problem can be computationally efficient
and intuitive~\cite{zhang1999,BarronM15}, but the limitations of the light and distribution
models applied to the data make them brittle in the presence of input noise,
which is unavoidable in real data.
Deep learning based methods have achieved impressive
results in reducing this brittleness in outdoor depth reconstruction for autonomous driving~\cite{fu2018}
and indoor geometry reconstruction~\cite{Eigen2014,Tulsiani2017}.

\textbf{Single-photo body estimation} methods typically bottleneck through fixed intermediate representations, which while enabling piecewise modeling, ultimately limit the amount of achievable detail. Some methods bottleneck through segmented images~\cite{moviereshape, hasler10multilin, tan17indirect, Dibra2017, omran2018neural}, others through estimated keypoints positions~\cite{bogo2016keep, lassner17unite}, and some through both~\cite{varol18bodynet,pavlakos18learning,Guo2012ClothedAN,alldieck19cvpr}. All such methods permit too much ambiguity to allow for dense surface reconstruction.  Recent methods~\cite{kanazawa2018end} avoid this limitation by using encoder-decoder representations directly on the image. They achieve remarkable robustness to images in the wild, but struggle to recover detailed shape and pose. Work on SURREAL~\cite{varol2017learning} estimates depth directly, but with coarse detail. The SiCloPe~\cite{siclope} system tolerates greater clothing variation than our system, but its geometric detail is limited by the use of intermediate silhouettes. To the credit of these works, all but~\cite{Dibra2017} were designed for capturing bodies ``in the wild'' with tolerance for pose variation, whereas our goal is to capture a detailed avatar from a restricted pose.

\textbf{Single-photo face estimation} methods have produced useful insights for body estimation. Early work by Blanz and Vetter~\cite{blanz99morphable} was ground-breaking but suffered from lack of detail and problems with robustness \emph{in the wild}. Robustness was addressed by data-driven models~\cite{booth17threedcvpr,dou17endcvpr, jackson17largeiccv,sengupta17sfsnet,tewari17mofaiccv,tewari18selfsupervised,tran17regressingrobust}; detail was addressed first by shape from shading~\cite{kemelmacher11threedtpam,li14intrinsiceccv}, and then by deep learning~\cite{sela2017unrestricted,tran2018extreme,richardson17learningcvpr}. A recent survey by Zollhoffer et al~\cite{zollhoffer18stateeccv} has more specifics. FAX specifically shares themes with~\cite{sela2017unrestricted,tran2018extreme}, in which the image-to-image translation architecture from Isola et al~\cite{isola2017image} is successfully applied to detailed face geometry estimation.

Our focus is on avatar geometry estimation from a single color image. For a more general review of body estimation from multiple images, readers are advised to review the excellent summaries of previous work provided in Alldieck et al~\cite{alldieck2018detailed} and Bogo et al~\cite{bogo17dynamicfaust}.

\section{Method}\label{method}

\begin{figure*}[htb]
    \centering
    \begin{overpic}[trim={0cm 12cm 0cm 0cm},clip,width=\linewidth]{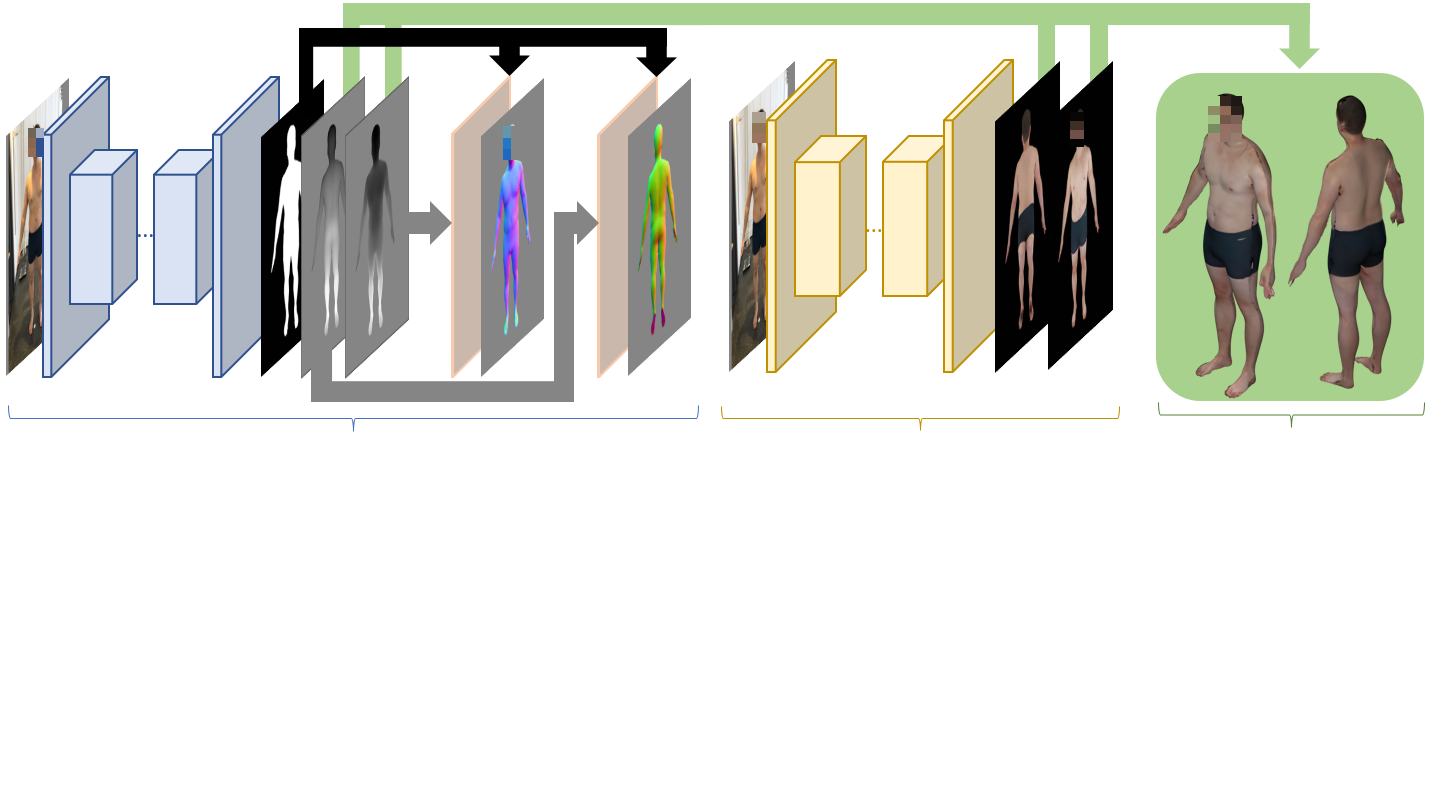}
      \put(1,8.4){{\small \color{white} a}}
      \put(18.8,8.4){{\small \color{white} b}}
      \put(21.8,8.4){{\small \color{white} c}}
      \put(24.8,8.4){{\small \color{white} d}}
      \put(31.7,8.4){{\small \color{white} $\delta$}}
      \put(33.8,8.4){{\small \color{white} e}}
      \put(41.9,8.4){{\small \color{white} $\delta$}}
      \put(44.0,8.4){{\small \color{white} f}}
      \put(51.0,8.4){{\small \color{white} a}}
      \put(69.5,8.4){{\small \color{white} g}}
      \put(73.0,8.4){{\small \color{white} h}}
      \put(58.3,1.1){{\small \color{Dandelion} albedo inference}}
      \put(85,1.1){{\small \color{OliveGreen} output mesh}}
      \put(18.5,1.1){{\small \color{Periwinkle} depth inference}}
    \end{overpic}
    \caption{Network architecture for geometry inference (left) and albedo inference (right) from an image a.
      They share the architecture inspired by~\cite{wang2017high} in the first stage (blue and yellow, trained separately).
      In geometry, the network outputs three channels (mask b, front and back depth e and d),
      while in albedo six channels are produced (RGB back g and RGB front h).
      Depth channels are processed by fixed spatial differentiation
      layers $\delta$ which use the mask to limit its effect to the foreground area,
      resulting in front and back normals (e and f). For compactness, we do not show the front and back albedo discriminators.
    }
    \label{fig:net}
\end{figure*}
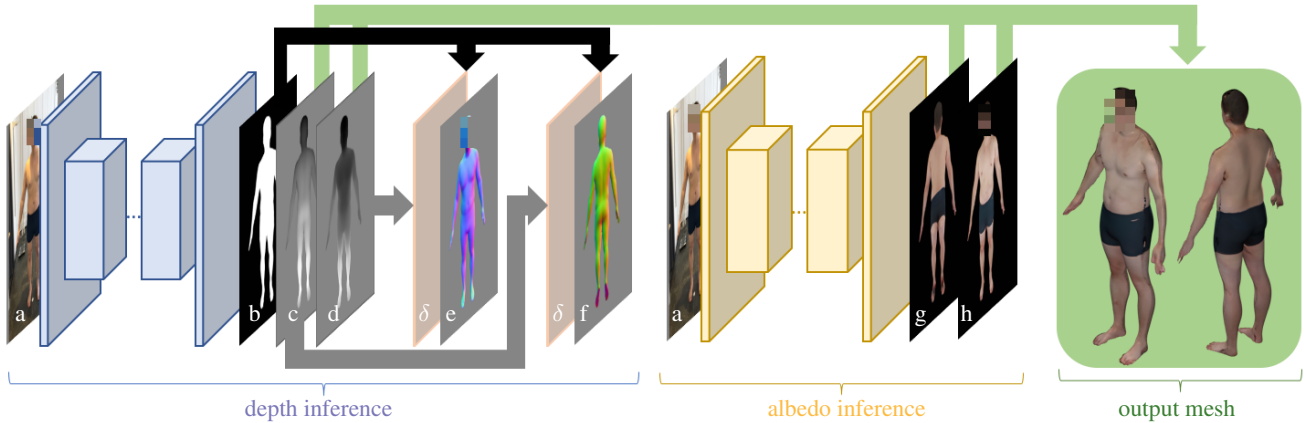

Our goal is to estimate a detailed 3D scan from a single RGB image.
We treat this as an image-to-image translation task, where we \emph{translate}
an image to depth and albedo values in image space.
More specifically, we estimate those outputs for both the front- and back-facing
portions of the body. The depth images form regular grids of vertices,
which can be trivially triangulated to create a 3D surface.

We describe our depth estimation architecture in more detail in Section~\ref{sec:depthestimation},
but focus first on albedo estimation in Section~\ref{sec:albedo}, since the training protocol
closely resembles the prior work of~\cite{wang2017high}. Finally, we explain how to
obtain a complete, reposable and reshapable avatar in Section~\ref{sec:alignment}.

\subsection{Albedo estimation}\label{sec:albedo}

Our architecture of choice is based on the image-to-image translation work
of~\cite{wang2017high}. We omit features specific to semantic segmentation and
image editing, as well as their ``enhancer'' networks. Thus we define our generator
using their ``global generator'', which is composed of a downsampling
section, followed by a number of residual~\cite{he2016deep} blocks, and completed with
an upsampling section that restores the feature maps to the input resolution. We make
one minor modification by replacing transposed convolutions with
upsample-convolutions to avoid checkerboard artifacts~\cite{odena2016deconvolution}.

The loss in~\cite{wang2017high} is composed of three terms: an adversarial loss,
$\mathcal{L}_{GAN}$ using a multi-scale PatchGAN~\cite{isola2017image}
discriminator with an LSGAN~\cite{mao2017least} objective; a feature matching loss,
$\mathcal{L}_{FM}$, which penalizes discrepancies between the internal discriminator
activations from the generated $G$ vs.\ real images $y$; and a perceptual loss,
$\mathcal{L}_{VGG}$, which uses a pre-trained VGG19 network, and similarly measures
the different VGG activations from real and generated images:
\begin{align}\label{eq:pix2pixHD_loss}
    \mathcal{L}_{alb}\left(x, y^i\right) &= \mathcal{L}_{GAN}\left(x, y^i, G^i\right)
    + \lambda_{FM}\mathcal{L}_{FM}\left(x, y^i, G^i\right) \nonumber \\
    &+ \lambda_{VGG}\mathcal{L}_{VGG}\left(y^i, G^i\right)
\end{align}
where $i$ indexes front and back. Every generated image $G^i(x)$ depends on the input image $x$,
so we drop this dependency from now on to simplify notation.
Front and back albedo use the same loss components,
though employ separate discriminators for the front and back estimates,
enabling them to specialize.
The application of this network to our problem of albedo
estimation is straightforward. Given synthetic training data (see Section~\ref{datasets})
of images and the corresponding front and back albedo, we estimate $G$ with six channels
corresponding to the two albedo sets (center of Figure~\ref{fig:net}).
The total loss is the sum of losses applied to front and back, $\mathcal{L}_{alb}\left(x, y^f\right) + \mathcal{L}_{alb}\left(x, y^b\right)$.

\subsection{Depth estimation}\label{sec:depthestimation}

\paragraph{Motivation}
As previously explained, direct estimation of depth is challenging due to
various reasons. First, there is an ambiguity between scale and distance to
the camera difficult to resolve even by humans. And second, this distance
to the camera entails a much larger data variance than shape details.
Therefore, a loss on depth encourages the network to solve the overall distance
to the camera, which is a very challenging and mostly irrelevant problem for our purpose.
Instead, we focus on inferring \emph{local} surface
geometry, which is invariant to scale ambiguities.

In initial experiments we managed to estimate detailed surface normals through the direct application
of the image-translation network described in Section~\ref{sec:albedo}.
However, integrating normals into robust depth \emph{efficiently}
is a challenging problem at the core of shape from shading literature.
While integration of inferred normal images is challenging and expensive,
its inverse operator is simple: the spatial derivative.
Spatial derivatives can be implemented simply as a fixed layer with a local difference filter.
By placing such layer directly behind the estimated normals (see $\delta$ layer in Figure~\ref{fig:net}),
we are implicitly forcing the previous result to correspond to depth.
Similar to the classic integration
approach, this allows us to infer depth even in the absence of depth ground
truth data, but without the extra computational cost incurred by explicit integration.

\paragraph{Losses}

In our depth architecture (see Figure~\ref{fig:net}), the output
is three channels and they represent the front and back depth $G_d^i$
where $i$ denotes front or back, as well as a mask $G_m$ denoting where depth is valid.
The front and back depth are processed
with a spatial differentiation network $\delta$ 
that
converts the depth into normals $G_n^i = \delta(G_d^i, G_m, f)$.
This spatial differentiation depends on the focal length $f$ (considered fixed in train and test
data) to correct perspective distortion. Furthermore, the differentiation operator
incorporates the mask $G_m$ produced by the network, to ensure we do not differentiate
through boundaries. In the areas where depth is not valid, a constant normal value is
produced.

While albedo (or color in general) seems to clearly benefit from adversarial losses,
the same does not seem to be true for recovering geometry. In our experience
(similar to what is described in~\cite{sela2017unrestricted}),
the adversarial loss in $\mathcal{L}_{alb}$ introduces noise when applied to the problem of depth and normal estimation,
and reduces its robustness to unseen conditions.
For this reason, the depth $\mathcal{L}_{d}$ and normal $\mathcal{L}_{n}$ terms of our geometry estimation objective

\begin{align}\label{eq:pix2pixHD_loss}
    \mathcal{L}_{n}^i\left(x, y\right) &= \mathcal{L}_{L1}\left(y, G_n^i\right) + \lambda_{VGG}\mathcal{L}_{VGG}\left(y, G_n^i\right) \\
    \mathcal{L}_{d}^i\left(x, y\right) &= \mathcal{L}_{L1}\left(y, G_d^i\right)
\end{align}
replace the adversarial loss with an L1 loss. $\mathcal{L}_{VGG}$ is not applied to
the depth representation as this would require a normalization of the (unbounded)
depth values that could cause training instability. The total loss can potentially
include this geometric loss applied to normals and/or depth, as well as a binary
cross entropy loss on the mask output

\begin{align}\label{eq:pix2pixHD_loss}
    \mathcal{L}_{full}\left(x, y\right) &= \lambda_{d}\left(\mathcal{L}_{d}^f + \mathcal{L}_{d}^b  \right)
    + \lambda_{n}\left(\mathcal{L}_{n}^f + \mathcal{L}_{n}^b \right) \nonumber \\
    &+ \lambda_{msk}\mathcal{L}_{msk}\left(y_m, G_m\right)
\end{align}

In Section~\ref{sec:abl_eval} and Table~\ref{fig:model_variants}, we study the
contributions of these loss terms both qualitatively and quantitatively.

\subsection{Estimating Dense Correspondence}\label{sec:alignment}

The system described in the previous section produces per-pixel depth values,
which are inherently incomplete. Moreover, since those values are created per pixel,
they lack any semantic meaning (where is the nose, elbow, etc).
In this section we adopt the mesh alignment process described in~\cite{Bogo2015detailed}
to infer the non-visible (black parts in Figure~\ref{fig:silhouette}) parts of the body
geometry based on SMPL~\cite{loper2015smpl}, a statistical model of human shape and pose.

The alignment process deforms a set of free body vertices (referred to as \emph{the mesh}) so that they are
close to the pointcloud inferred in the previous section (referred to as \emph{the scan}),
while also being likely according to the SMPL body model.
Similar to \cite{Bogo2015detailed}, we minimize a loss
composed of a weighted average of a scan-to-mesh distance term $E_{s}$,
a face landmark term $E_{face}$, two pose and shape priors $E_{pose}$ and $E_{shape}$,
and a term that couples the inferred free vertices with the model $E_{cpl}$.
We provide some intuition about the terms in the following paragraphs,
although the details can be obtained in the original publication.

$E_{s}$ penalizes the squared 3D distance between the scan
and closest points on the surface of the mesh.
$E_{face}$ penalizes the squared 3D distance between detected face landmarks~\cite{KazemiS14}
on the image (in implicit correspondence with the scan) and pre-defined landmark locations in SMPL.
$E_{cpl}$ encourages the mesh, which can deform freely, to stay close
to the model implied by the optimized pose and shape parameters.
$E_{pose}$ and $E_{shape}$ regularize pose and shape of the coupled model by penalizing the Mahalanobis
distance between those SMPL parameters and their Gaussian distributions inferred
from the CMU and SMPL datasets~\cite{bogo2016keep}.

\begin{figure*}[t]
    \centering
    \includegraphics[trim={0 8.5cm 0 8.5cm},clip,width=\linewidth]{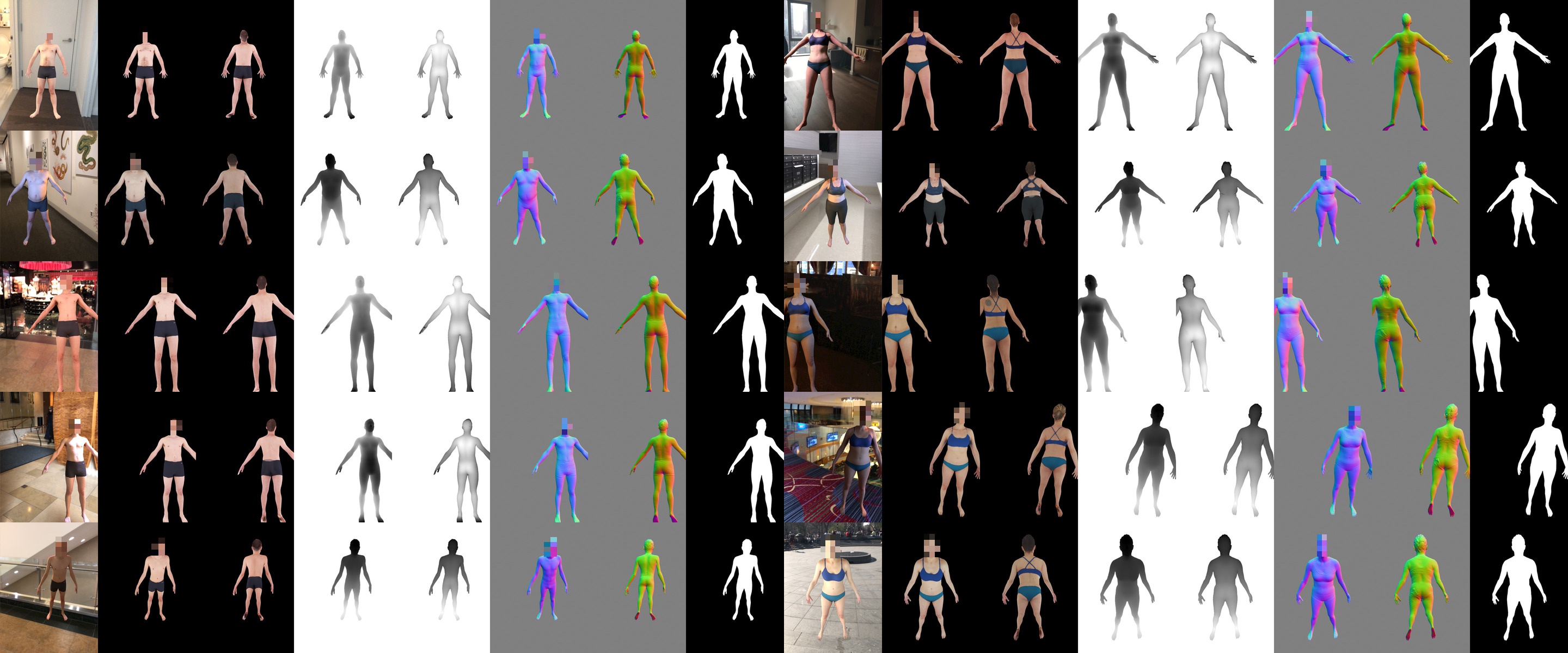}
    \caption{{
        Each row shows two instances of synthetic data (one male, one female).
        For each example, left to right: RGB, front and back albedo, front and back depth,
        front and back normals, and segmentation. Note that these examples do not really
        belong to our training set, since the textures come only from test subjects who signed
        a special consent form.
      }
    }
    \label{fig:synth}
\end{figure*}

As it is common in single view and non-calibrated multi-view shape estimation, our
results cannot recover the subjects scale accurately.
Since SMPL cannot fit scan at arbitrary scales,
we first scale the scan to a fixed height before
optimizing the mesh, then apply the inverse scale to the optimized mesh, returning it
to the original reference frame.

When training our depth estimator, the loss on depth acts as a global constraint,
enforcing that the front and back scans be estimated at consistent scales. When this
loss is omitted during training (see Section~\ref{sec:abl_eval}),
the front and back scale are not necessarily coherent,
and thus their relative scale must be optimized during mesh alignment. This can be accomplished
by introducing a single additional free scale variable that is applied to the back vertices
and optimized along with the mesh. When describing our experiments, we refer to this
option as \emph{opt back}.

\section{Experiments}\label{experiments}

\subsection{Training and evaluation details}\label{training_details}

For albedo estimation, we train on random crops of size $512\times512$ to
comply with memory limitations. The multi-scale discriminators process images at $1\times$,
$\frac{1}{2}\times$, and $\frac{1}{4}\times$ resolutions. Losses are weighted as
in~\cite{wang2017high}. For depth estimation, we train on $720\times960$
images, and work with a focal length of 720 pixels. We do not assume a fixed
distance to the camera. Both albedo and depth estimation networks are trained for 180k steps with a
batch size of one, and input images are augmented with gaussian
blur, gaussian noise, hue, saturation, brightness, and contrast. The training process
takes approximately 48 hours with a V100 Tesla GPU.

Evaluation is performed on $720\times960$ images. A single forward pass of either
network takes about 100 milliseconds, while aligning SMPL to the scan
takes 7 seconds.

\subsection{Datasets}\label{datasets}

We train exclusively on synthetic datasets~(Figure~\ref{fig:synth}), and test
on real images collected ``in-lab'' --- i.e., in a well-lit, indoor
environment, where images are captured by lab technicians, and subjects wear
tight-fitting clothing and stand in an ``A''-pose (see
Figure~\ref{fig:seven_examples}).

We render 40,000 synthetic image tuples (1\% held out each for validation and
testing). The bodies have a base low-frequency geometry synthesized with SMPL,
and high-frequency displacements captured in-lab.
The SMPL shape parameters are sampled from the CAESAR
dataset and poses are sampled from a mix of (a) CAESAR poses and (b) a set of
in-lab scan poses with arms varying from A-pose to relaxed.
Textures and displacement maps, derived
from 3D photogrammetry scans of people captured in-lab, are randomly sampled and applied
to the base bodies, which increases the diversity of the input and output spaces.

The camera is
fixed with zero rotation at the origin, and the body randomly translated and
rotated to simulate a distance of roughly 2 meters with a slight downward tilt
of the camera. Specifically, translation is sampled from $x \sim [-0.5, 0.5],
y \sim [0.0, 0.4], z \sim [-2.2, -1.5]$ in meters and rotation as Euler angles
in degrees from $x \sim [-9.0, 35], y \sim [-7, 7], z \sim [-2, 2]$, applied
in $yxz$ order.
Background images are drawn from OpenImages~\cite{krasin2016openimages},
excluding images containing people.

We use three light sources: an image-based environment light (which uses the
background image as a light source), a point light, and a rectangular area
light. For each render, we randomly sample the intensity of all lights, the
position and color temperature of the point and area lights, the orientation
and size of the area light, and the specularity and roughness of the shader on
the body. All light sources cast raytrace shadows, with the most visible
generally coming from the area and point lights.

\begin{table}[htb]
    \centering \small
    \rowcolors{2}{gray!25}{white}
    \begin{tabular}{cccc}
        \midrule
        Subject ID & FAX (mm) & FAX (mm) (opt pose) & \cite{alldieck2018video} \\
        \midrule
        50002 & 9.46 & 6.56 & 5.13\\
        50004 & 7.90 & 4.19 & 4.36\\
        50009 & 5.23 & 3.86 & 3.72\\
        50020 & 6.60 & 3.85 & 3.32\\
        50021 & 4.76 & 3.27 & 4.45\\
        50022 & 5.08 & 3.50 & 5.71\\
        50025 & 5.03 & 3.02 & 4.84\\
        50026 & 7.83 & 4.87 & 4.56\\
        50027 & 8.21 & 4.34 & 3.89\\
        \bottomrule
    \end{tabular}
    \caption{Bi-directional mesh-to-mesh error on subjects from D-FAUST dataset using our baseline method. For each subject we report average error across multiple instances rendered with
    random environment configurations, using the methodology described in Section~\ref{datasets}.}
    \label{fig:d_faust}
    \vspace{-2mm}
\end{table}
\subsection{Visual Evaluation}\label{visual_evaluation}

As a baseline, we consider direct estimation of frontal depth with an $L1$
loss function. Figure~\ref{fig:d_l1_vs_n_l1} shows meshes estimated from
natural test images, comparing models trained with an $L1$ loss on depth vs.\
an $L1$ loss on normals. Results with the depth-only loss appear unusable,
while results with the normals-only loss are smooth, robust, and capture an
impressive amount of detail. Thus, for detailed
depth estimation of human bodies, a direct loss on depth is insufficient,
whereas a loss on surface normals is sufficient to produce robust and detailed
depth estimates. However, since the loss on normals only constrains the output locally,
the geometry will not be true to scale. A loss on depth, while not crucial for the
quality of the geometry, encourages the output toward a space of plausible human
scales.

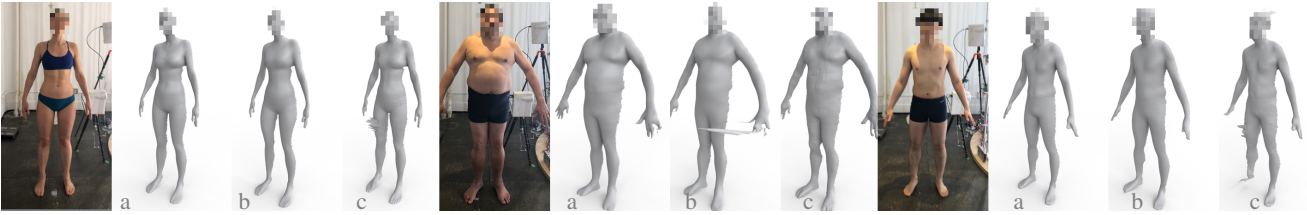
\begin{figure*}[t]
    \centering
    \begin{overpic}[trim={0 0cm 0 1cm},clip,width=\linewidth]{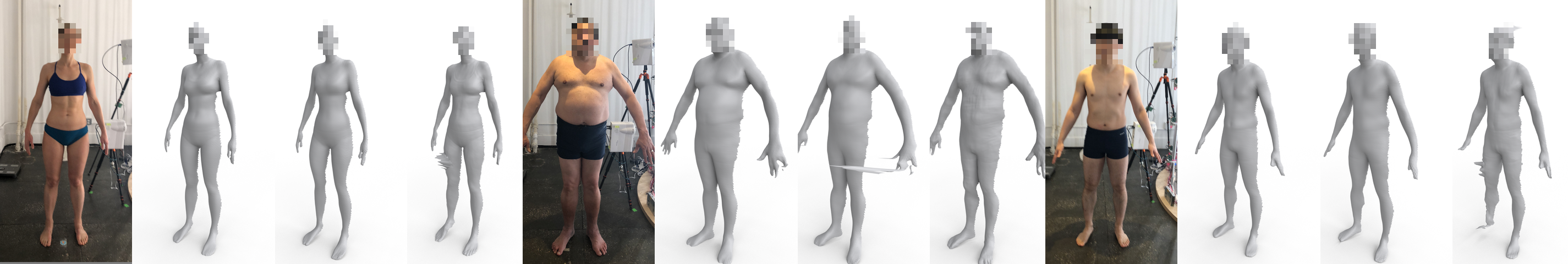}
      \put(9,.1){{\small \color{gray} a}}
      \put(18,.1){{\small \color{gray} b}}
      \put(27,.1){{\small \color{gray} c}}
      \put(43,.1){{\small \color{gray} a}}
      \put(52,.1){{\small \color{gray} b}}
      \put(61,.1){{\small \color{gray} c}}
      \put(77,.1){{\small \color{gray} a}}
      \put(86,.1){{\small \color{gray} b}}
      \put(95,.1){{\small \color{gray} c}}
    \end{overpic}
    \caption{Visual comparison of the ablation experiments (a) Baseline, (b) without $\mathcal{L}_{L1}\left(y, G_n^i\right)$ term, and (c) 2 scales.}
    \label{fig:vis_abl}
\end{figure*}

\begin{figure*}[h!]
    \centering
    \includegraphics[trim={0 0cm 0 0},clip, width=\linewidth]{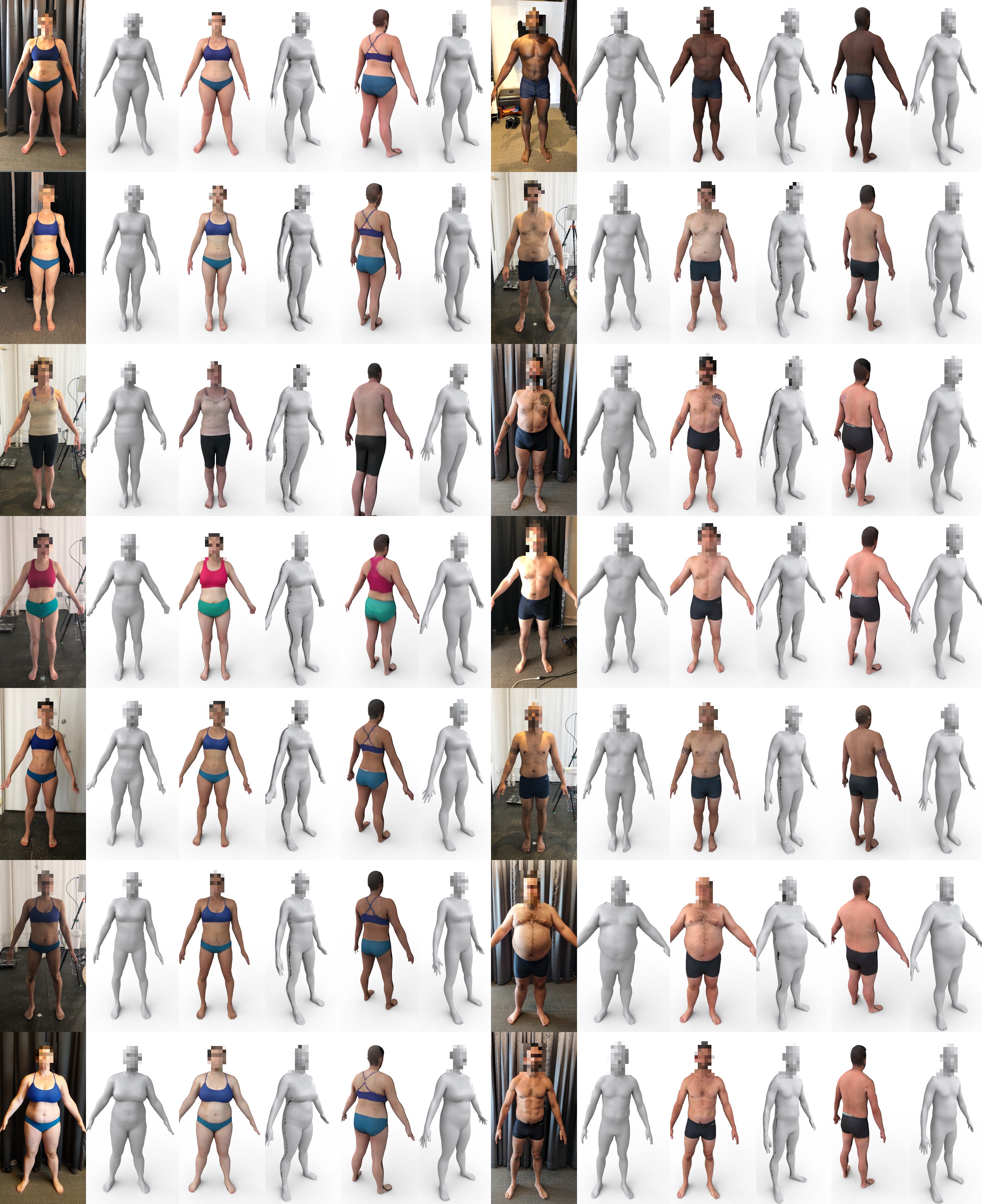}
    \caption{Two columns with RGB image, scan with and without texture and alignment. Pay close attention to variation
      in shape, pose and ethnicity, as well as the fidelity of detail in hips, waist and chest, specially in the silhouette region.
      Note that most test subjects in this figure are wearing similar clothes to the garments present in the synthetic training data.
    }
    \label{fig:seven_examples}
\end{figure*}

One advantage of FAX is its ability to extract subtle shape detail from a
single image. Recovered shapes are intricate and personal, as observed in
the waist, hips and chest of almost every example in
Figure~\ref{fig:seven_examples}. This is hard to achieve by methods based on
convex-hull~\cite{siclope}, voxels~\cite{varol18bodynet} or SMPL shape
parameters~\cite{kanazawa2018end}. Even methods optimizing explicitly the
shape to fit the image contour, like~\cite{alldieck2018video}, fail to recover
this level of detail because the underlying optimization has to find a
compromise between the data and the underlying (overly smooth) model.  Detail
obtained from FAX is mostly visible in the contours, but the side renders show
that this detail is reconstructed in a coherent manner across the body shape,
recreating bust and stomach shape that is coherent with the silhouette and
image shading.

Visual discontinuities such as shadows and tattoos are a challenge. Classic
shape-from-shading methods are notorious for introducing ridge artifacts at
misleading visual boundaries. As shown in Figure~\ref{fig:seven_examples} (row
3, on right), our methods produce clean geometry in the presence of tattoos.
And in Figure~\ref{fig:vis_abl}, our method exhibits invariance to sharp
shadows. We credit this invariance almost entirely to the diversity in our
training dataset; before introducing sharp shadows in our training
(Figure~\ref{fig:synth}: row 3 on left), ridge artifacts around shadows were
common in our test output.

Spatial scan holes are an additional challenge. Like many high-quality scanner
setups, our raw estimated scan does not capture all geometry, noticeably
visible as the seam between front and rear-facing depth maps. This problem is
one motivation for fitting an avatar: beyond providing reposability, it
provides hole closure and scan completion. Figures~\ref{fig:teaser}
and~\ref{fig:seven_examples} illustrates our scans, their seams, and the
avatars that provide hole closure.

Our front albedo estimation network is resilient to soft shadows. To see this,
consider the RGB input and frontal textured scan in
Figure~\ref{fig:seven_examples}, which is iluminated with the same light as
the grey scans. In particular, observe the removal of skin highlights in row 4
right, and much more even skin tone in legs and torso in most of them, e.g.
row 7 right. Removing sharp casted shadows is extremely challenging, but
reasonable results are achieved in row 1, 2 and 5 right.

Our back albedo estimator exhibits pleasing front/back consistency, including
skin tone and garment continuity. Some bra straps (e.g. row 7 left
in~\ref{fig:seven_examples}) show a continuous but physically implausible
configuration, while garments in skin-tone colors (row 3 left
in~\ref{fig:seven_examples}) blend into the skin texture. Improvements to
training data should address this.

\subsection{Quantitative evaluation on Dynamic FAUST}\label{sec:d_faust_eval}

We compare our system quantitatively with~\cite{alldieck2018video}, which is one of
the state of the art systems in estimating shape from multiple images. Following~\cite{alldieck2018video},
we generate synthetic renders from the subjects in Dynamic FAUST, estimate their
shape, and evaluate it against the synthetic data. Unlike~\cite{alldieck2018video},
we only require one image for each subject. We should also note that since our
system works with RGB images, the authors of~\cite{bogo17dynamicfaust} kindly
provided us with one natural texture for each subject in their dataset.

We follow the procedure described in~\cite{alldieck2018video} to compute the errors
in Table~\ref{fig:d_faust}. First, we estimate the scan and alignment as described
in Sections~\ref{sec:depthestimation} and \ref{sec:alignment}.
Using SMPL, we unpose the alignment and scale it to make it as
tall as the groundtruth shape. Using this fixed shape, we optimize translation and
scale to minimize the average bidirectional distance between vertices in each mesh and the
surface of the other mesh, initializing the translation and pose
from groundtruth. We repeat this procedure over N synthetic images per subject to obtain more
reliable estimations of the error.
This average bidirectional distance is reported in the left column from Table~\ref{fig:d_faust}.
This procedure is comparable to the \emph{full method} reported in ~\cite{alldieck2018video}.
Our errors are larger than in~\cite{alldieck2018video}, which can be attributed
to two factors. First, we have access to a single image while~\cite{alldieck2018video}
used hundreds of them. Second, applying the groundtruth pose from the scan can be suboptimal,
since SMPL conflates pose and shape to some extent. To decouple this problem, we
also optimized the pose together with scale and translation (keeping shape fixed at all times),
which is shown in the middle column of Table~\ref{fig:d_faust}. Note however
that we believe this result is not directly comparable to~\cite{alldieck2018video}.

\sisetup{detect-weight,mode=text}
\renewrobustcmd{\bfseries}{\fontseries{b}\selectfont}
\renewrobustcmd{\boldmath}{}
\newrobustcmd{\B}{\bfseries}

\newrobustcmd{\normalsterm}{$\mathcal{L}_{n}^i$}
\newrobustcmd{\depthterm}{$\mathcal{L}_{d}^i$}
\newrobustcmd{\vggterm}{$\mathcal{L}_{VGG}$}

\begin{table*}[h]
    \centering \small
\rowcolors{2}{gray!25}{white}
\begin{tabular}{rccccc|ccccccc}
\toprule
  &  &  & Blur & \# Res.  &  & Error  & Error  & Error   & Error   \\
Label & \normalsterm & \depthterm & aug. & blocks  & \# scales &  & (opt back) &  (opt pose) &  (opt back, pose)  \\
\midrule

Baseline       & \cmark & \cmark & \cmark  & 9 & 4 & 6.89  & 6.66  & 3.77 & 3.65    \\

5 res blocks   & \cmark & \cmark & \cmark  & 5 & 4 & 6.76  &  6.63 & 3.62  & 3.60 \\

No blur aug.   & \cmark & \cmark & \xmark  & 9 & 4 & 6.99  & 6.97  & 3.83  & 3.85 \\

2 scales       & \cmark & \cmark & \cmark  & 9 & 2 & 8.21  &  7.88 & 4.50  & 4.34 \\

No depth      & \cmark & \xmark & \cmark  & 9 & 4 &  -    & 8.57  & -       & 3.87 \\

No normals & \xmark & \cmark & \cmark  & 9 & 4 & 9.02  & 9.04  & 5.28 & 5.36 \\

No VGG & \cmark & \cmark & \cmark & 9 & 4 & 7.80 & 6.69 & 4.18 & 3.60 \\

\bottomrule

\end{tabular}
\caption{Ablation study on our depth estimator, using mesh distance for evaluation. See Section~\ref{sec:abl_eval} for more details.}
\label{fig:model_variants}
\end{table*}

\subsection{Ablation Study}\label{sec:abl_eval}

Here we study factors that contribute to our method performance.
We first consider the individual contribution of our loss terms.
We next vary the number of residual blocks in the network, which affects network depth.
Similarly, we change how many downsampling operations (\emph{scales})
are performed. These operations involve learned convolutions,
and thus add capacity and depth to the network. Finally, we test the role
of blur data augmentation performed on our synthetic training data.
We run this experiment on images from 87 subjects (see Figure~\ref{fig:vis_abl} for
four subject examples).

Results of the ablation study are summarized in
Table~\ref{fig:model_variants}. For compatibility with~\cite{alldieck2018video}, we perform all
comparisons with estimated alignments instead of scans, using the procedure described in
Section~\ref{sec:d_faust_eval}, reporting average bi-directional point-to-mesh
distances. However, fitting a model to our scan regularizes problems in less robust
variants of our pipeline (e.g., ``No blur aug.'') and the imperfections in the
unposing process may introduce subtle and potentially misleading inaccuracies,
thus the tradeoffs in model variants will not necessarily be well represented by
this metric.

Columns labeled with \emph{opt pose} relate to pose optimized to
minimize distance, similar to the previous section. We also consider the
independent optimization of front and back scale (as described in
Section~\ref{sec:alignment}, labeled as \emph{opt back}), since experiments
with no depth show differences in scale in the front and back that render
quantitative evaluation useless without such independent optimization.

Most noticeable is the importance of normals in this loss. Removing
normal terms (both L1 and VGG) is more detrimental than removing the depth term, which
is consistent with the intuition provided in Figure~\ref{fig:d_l1_vs_n_l1}.
Removing depth or normal terms incurs a negative effect compared with the
baseline. Reducing downsampling makes the network shallower, allowing it to
keep more detail (see Table~\ref{fig:model_variants}) but also noise,
incurring a big accuracy penalty. Although blur augmentation has a small
numerical impact, we observe that it creates spikes and holes,
making it unusable for the rapid creation of a textured scan. Lastly,
omitting the VGG loss on normals causes a minor loss in accuracy.

We add an extra configuration in Figure~\ref{fig:vis_abl}: removing the L1
loss on normals but keeping VGG results in an oversmoothed scan with more
shading artifacts. Finally, while it's surprising that reducing the number of
residual blocks improves accuracy, we consider the difference negligible.

\section{Conclusions}

FAX estimates full body geometry and albedo from a single RGB image at a level of detail previously unseen.
This quality depends critically on two main factors. First, we do not indirect our
output through representations like voxels, convex hulls or body models,
which allow us to recover detail at the original pixel definition with an image-translation network,
orders of magnitude faster than competing methods.
Second, our geometry estimation depends critically on the role of surface normals,
and we show how even surface normals alone can produce plausible bodies in the absence of depth information.
We evaluate our system using two datasets, perform an ablation study,
and extensively illustrate the visual performance of our system.

For future work, we believe improving our training data can overcome many
restrictions of the current method, like the frontal pose or minimal clothing.
We would like to eliminate the seams in scan geometry and texture in a rapid,
data-driven manner. Finally, we believe incorporating an additional view
can help reduce the inherent ambiguity present in the shapes
estimated from a single view.

\clearpage
\newpage

\clearpage
\newpage

{\small
\bibliographystyle{ieee_fullname}
\balance
\bibliography{egbib}
}

\end{document}